\documentclass[%
 preprint,
 amsmath,amssymb,
 aps, physrev,
]{revtex4-2}

\usepackage{graphicx}% Include figure files
\usepackage{dcolumn}% Align table columns on decimal point
\usepackage{bm}% bold math
\usepackage{siunitx}
\usepackage{booktabs}
\usepackage{mhchem}
\usepackage{hyperref}
\begin{document}

\preprint{APS/123-QED}

\title{\textbf{Thermodynamically Consistent Neural Networks for a Universal Fluid Gibbs Free Energy Equation of State} 
}% 

\author{Xurui Li}
\thanks{These authors contributed equally to this work.}
\affiliation{%
 Department of Chemical Engineering, Tsinghua University, Beijing 100084, People's Republic of China
}%
\author{Zhiguo Gan}
\thanks{These authors contributed equally to this work.}
\affiliation{%
 Department of Chemical Engineering, Tsinghua University, Beijing 100084, People's Republic of China
}%
\author{Jiaming Zhang}
\thanks{These authors contributed equally to this work.}
\affiliation{%
 Department of Chemical Engineering, Tsinghua University, Beijing 100084, People's Republic of China
}%
\author{Hongxi Zeng}
\affiliation{%
  School of Engineering, The University of Edinburgh, King's Buildings, Edinburgh EH9 3DW, United Kingdom
}
\author{Diannan Lu}%
 \email{Contact author: ludiannan@tsinghua.edu.cn}
\affiliation{%
 Department of Chemical Engineering, Tsinghua University, Beijing 100084, People's Republic of China
}%

% \collaboration{CLEO Collaboration}%\noaffiliation

\date{\today}% It is always \today, today,

\begin{abstract}
Over the past three centuries, many thermodynamics researchers have worked to develop fluid equations of state with both high accuracy and broad applicability, proposing various theoretical frameworks and models.
Building on these efforts, this study uses Gibbs free energy as the starting point and introduces thermodynamically consistent neural networks to construct a fluid equation of state based on Gibbs free energy (GEoS). This approach establishes a novel machine learning framework that provides a unified representation of fluid states.
GEoS performs well on the vapor- and liquid-phase density test sets, with average absolute relative deviations (AARDs) of 5.25\% and 3.96\%, respectively. It outperforms conventional volume-translated cubic equations of state and existing machine learning models, demonstrating its effectiveness. Further analysis using the UMAP dimensionality reduction method shows that most molecules in the test set belong to extrapolation systems, confirming the strong generalization ability of the model.
Finally, the test results show that the PC-SAFT equation fitted using GEoS maintains high predictive accuracy, providing a reliable basis for its practical application in industry.
\end{abstract}

\maketitle

\section{Introduction}
An Equation of State (EoS) relates state variables to describe thermodynamic properties under various conditions \cite{Callen1985,Eliezer1986,smith2021introduction,elliott2023properties}. Fluid EoS forms the basis for calculating various fundamental physical properties \cite{Georgios2020,Privat2023}. In the process industry, EoS is essential to determine states of matter and phase behavior to guide process design. For example, core thermodynamic calculations are closely linked to EoS in fields such as petroleum and petrochemical technologies \cite{DONG2014839,Kontogeorgis2021}, supercritical fluid technologies \cite{Marr2000,Sodeifian2024,Dashtian2024}, and CCUS technologies \cite{Rivas2016,Georgios2020,Ahmadi2021,Nazeri2021}.

Over past three centuries, researchers have proposed numerous EoS for pure fluids\cite{Georgios2020}. 
The most classic is cubic EoS\cite{Valderrama2003,Privat2023}, which remains a workhorse in many petroleum industry applications\cite{Kontogeorgis2021}. 
In 1873, van der Waals proposed the first equation to describe gas and liquid phases in a unified form\cite{smith2021introduction}. It serves as the foundation for cubic EoS\cite{Valderrama2003}, followed by models such as the Redlich-Kwong equation\cite{Redlich1949}, Soave-Redlich-Kwong equation\cite{SOAVE19721197}, and Peng-Robinson equation\cite{Peng1976}. 
However, these equations cannot accurately predict liquid density and near-critical behavior. To address this, researchers introduce various modifications\cite{Privat2023}, such as new $\alpha$-function methods\cite{TWU1991,PINAMARTINEZ2019} and volume translation\cite{PENELOUX1982,TSAI1998,LeGuennec2016}. 
Despite these corrections, cubic EoS still struggle to accurately describe the behavior of polar fluids and high-pressure liquids.

Since the cubic equation era, fluid EoS research has shifted from directly constructing $p$-$V$-$T$ relations to Helmholtz energy frameworks that derive EoS via partial derivatives, now the dominant approach. High-accuracy models built on this framework include the empirical Helmholtz energy equation\cite{LEMMON-RP10}, SAFT\cite{Chapman1989,Huang1990,Economou2002} and its derivatives such as CPA\cite{Kontogeorgis1996} and PC-SAFT\cite{Gross2001}. These are widely applied: REFPROP integrates the empirical Helmholtz energy EoS\cite{NISTREFPROP2022} as the highest accuracy benchmark in industry and science, and the international water‑steam standard IAPWS-IF97\cite{Wagner2000} follows the same paradigm. 
However, the parameters of these EoS are usually obtained by fitting the experimental data\cite{Bell2016,Winter2025}. When experimental data are unavailable, although group-contribution estimates\cite{Haslam2020} can be used, this alternative method has limitations in accuracy and applicability \cite{Winter2025}.

With the rapid development of machine learning (ML), neural networks provide a promising approach to advancing fluid equations of state (EoS), enabling the rapid development of highly accurate models\cite{Succetti2025}. 
In this area, some studies focus on energy-based EoS modeling. Chaparro et al. employ ML to predict the EoS of Mie model fluids\cite{Chaparro2023,Chaparro2024}, demonstrating the feasibility of energy-based EoS modeling with ML. However, model fluids behave differently from real fluids, so these models cannot be directly applied to predicting the properties of real fluids.
Other studies focus on predicting EoS parameters. Felton et al. propose ML-SAFT, a model that predicts PC-SAFT parameters using ECFP4 fingerprints\cite{Felton2024}. Winter et al. propose SPT-PC-SAFT, which uses a Transformer architecture to extract molecular features from SMILES strings and predict PC-SAFT parameters\cite{Winter2025}.
However, ECFP4 fingerprints mainly describe two-dimensional molecular structures and may not fully represent molecular conformations. In contrast, feature extraction from SMILES using Transformers is a highly black-box process. Both limitations reduce confidence in the extrapolation ability of these models.

\begin{figure}
    \centering
    \includegraphics[width=\linewidth]{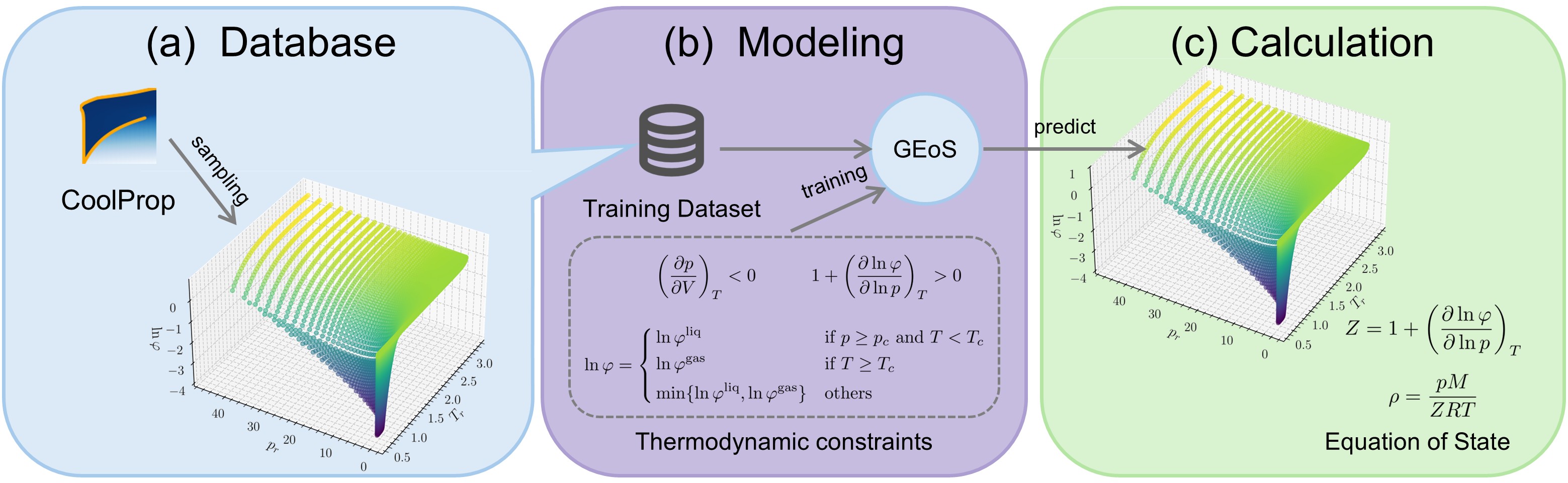}
    \caption{
        (a) Database: A database of fugacity coefficients for 101 substances is established based on CoolProp.
        (b) Modeling: Thermodynamic constraints, such as the mechanical stability condition and the non-negativity of the compressibility factor, are incorporated into the model.
        (c) Prediction: The model predicts the logarithm of the fugacity coefficient at a given temperature and pressure, from which other thermodynamic properties, such as the compressibility factor, are calculated via automatic differentiation.
    }
    \label{fig:TOC}
\end{figure}

Unlike existing studies that directly predict EoS parameters or residual Helmholtz free energy, this study adopts the Thermodynamically Consistent Neural Networks framework proposed in our previous work \cite{Li2026VP}. Molecular descriptors and the corresponding state variables are used as inputs to construct a predictive model for the residual Gibbs free energy of fluids. Based on this model, properties such as the compressibility factor are obtained through automatic differentiation. FIG.~\ref{fig:TOC} illustrates the workflow of this study.

\section{Method}
\subsection{Database}
The fugacity coefficient database is calculated by CoolProp\cite{Bell2014}, which is an open-source database of fluid properties, formulated based on the most accurate formulations in open literature. From the 124 fluids available in CoolProp, we select 104 substances that also appear in the DIPPR database, excluding predefined mixtures and rare gases.
In total, 10 atoms are considered: C, H, D, O, N, S, Si, F, Cl, and I.
Three substances are reserved for leave-one-out testing to facilitate parameter tuning and ablation experiments.
For more details, please refer to \textit{Section 1} of the \textit{Supporting Information}.

\begin{figure}
    \centering
    \includegraphics[width=0.5\linewidth]{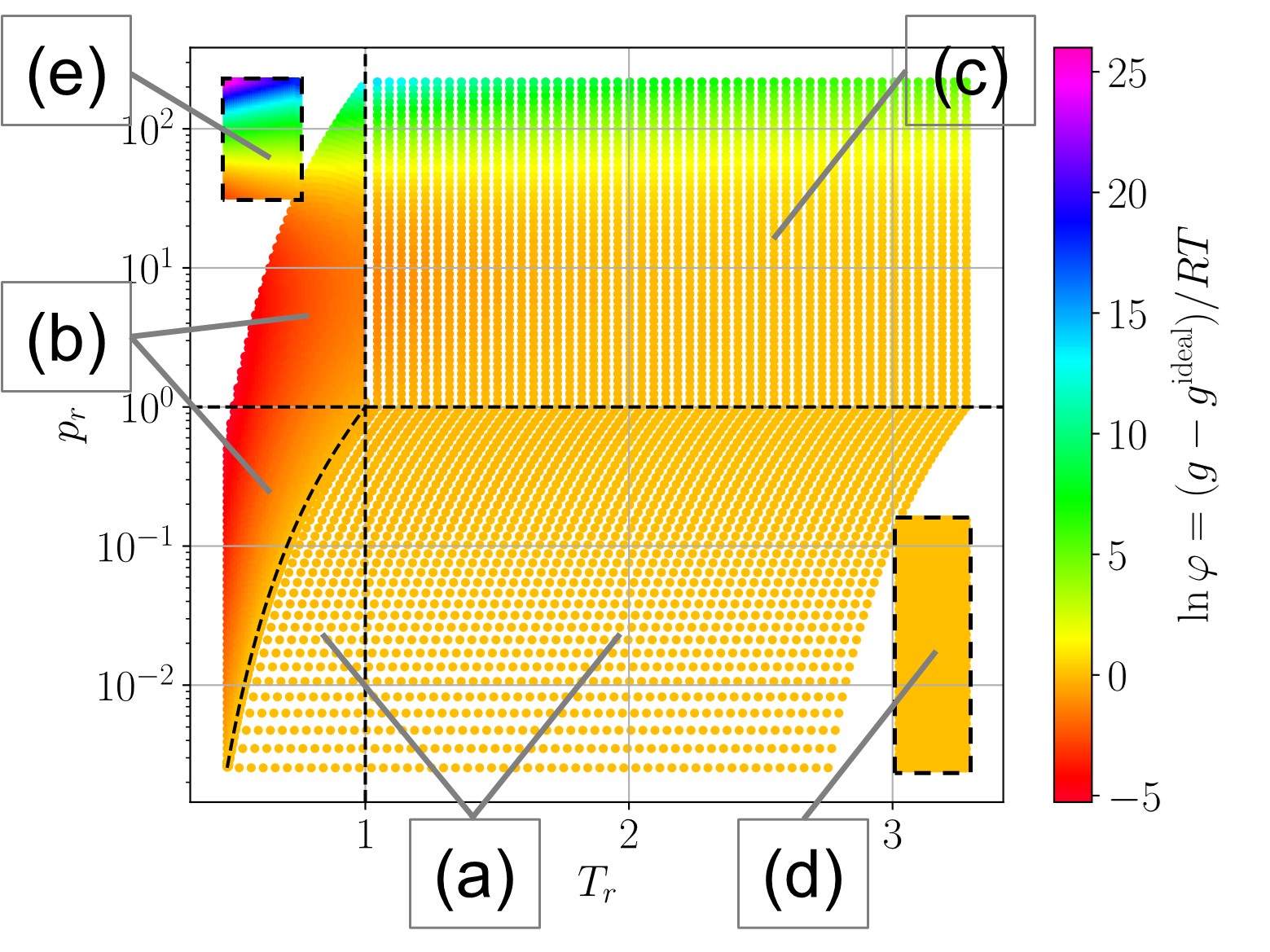}
    \caption{
        Sampling results of methane, colored by $\ln \varphi$.
        (a) The gas-phase region, where 2000 points are sampled;
        (b) The liquid-phase region, where 2000 points are sampled;
        (a, b) The intersection of the two regions, where an additional 2000 points are densely sampled near the vapor-pressure curve.
        (c) The supercritical region, where 2500 points are sampled;
        (d) An additionally supplemented low-pressure and high-temperature region, where 1600 points are sampled;
        (e) An additionally supplemented high-pressure and low-temperature region, where 1600 points are sampled.
    }
    \label{fig:Methane_Sample}
\end{figure}

The region-based sampling results are shown in FIG.~\ref{fig:Methane_Sample}. 2000 points are uniformly sampled by shifting the saturation line along the $T_r$ direction, and another 2000 points along the logarithmic $p_r$ ($\ln p_r$) direction. In the supercritical region, 2500 points are uniformly sampled over the $(T_r,\ln p_r)$ space. Due to the discontinuous derivative of the fugacity coefficient across the gas-liquid saturation line, 2000 points are densely sampled near the saturated vapor pressure curve.
To ensure that the model captures the behaviors at low-temperature ultra-high-pressure and high-temperature ultra-low-pressure, 1,600 points are sampled in each region. Specifically, due to the unique behaviors in these two regions, half of the sampled points are used as a validation set during training to serve as supervisory signals for model saving.
For more details, please refer to \textit{Section 1.2} of the \textit{Supporting Information}.

\subsection{Feature Engineering}
The molecular features of GEoS follow our previous strategy \cite{Li2026CP,Li2026VP}. 
RDKit \cite{RDKit2025} and Mordred \cite{Moriwaki2018} calculate 2D molecular descriptors, while PubChemFP \cite{NIH2025PubChem} serves as the 2D structural fingerprint. To supplement 3D structural information, Gaussian \cite{Gaussian} performs conformational optimization at the B3LYP/6-31G* level, yielding 25 quantum chemical descriptors. 
Previous studies demonstrate that models supplemented with quantum chemical descriptors can effectively differentiate isomers \cite{Li2026CP,Li2026VP}.

On this basis, with the critical molar density $\rho_c (\unit{kmol/m^3})$ as the prediction target, the Boruta algorithm \cite{JSSv036i11} is applied for feature selection in the entire DIPPR dataset, and features with a median importance rank not greater than two are selected as target features.
The feasibility of the Boruta selection scheme has been verified in our previous studies \cite{Li2026CP,Li2026VP}.
The purpose of performing selection on the entire DIPPR dataset is to make the feature space associated with our feature set more extensive because there are only 101 molecules available for training in Coolprop, and the feature set obtained by selection on it is not representative enough.

For more details on feature engineering design, please refer to \textit{Section 2} of the \textit{Supporting Information}.

\subsection{Gibbs Free Energy Thermodynamics}
Based on the model design, the Gibbs free energy of a pure substance is expressed as 
\begin{equation}
    \mu=\mu^\ominus+RT\ln{\left( p\varphi \right)} .
\end{equation}

According to the fundamental thermodynamic relation, the compressibility factor ($Z$) can be calculated
\begin{equation}
    \begin{aligned}
        Z = \frac{p V}{R T} &= 1 + p\left(\frac{\partial \ln \varphi}{\partial p}\right)_T \\
          &= 1 + \left(\frac{\partial \ln \varphi}{\partial \ln p}\right)_T
    \end{aligned}
    .
\end{equation}

According to the mechanical stability criteria, for a fluid to remain stable in a single phase, the following must be satisfied:
\begin{equation}
    \left(\frac{\partial p}{\partial V}\right)_T < 0 .
\end{equation}
This requires that a single-phase fluid should maintain a tendency for p to increase as $V$ decreases. Substituting this into $V = ZRT/p$, we can simplify to obtain:
\begin{equation}
    p\left(\frac{\partial Z}{\partial p}\right)_T < Z .
\end{equation}

The aforementioned mechanical stability criteria are embedded as prior physical constraints into the training of the model to ensure the thermodynamically consistent of the model's predictions.

\subsection{Model Design}
\subsubsection{Model Structure}
\begin{figure}
    \centering
    \includegraphics[width=0.6\linewidth]{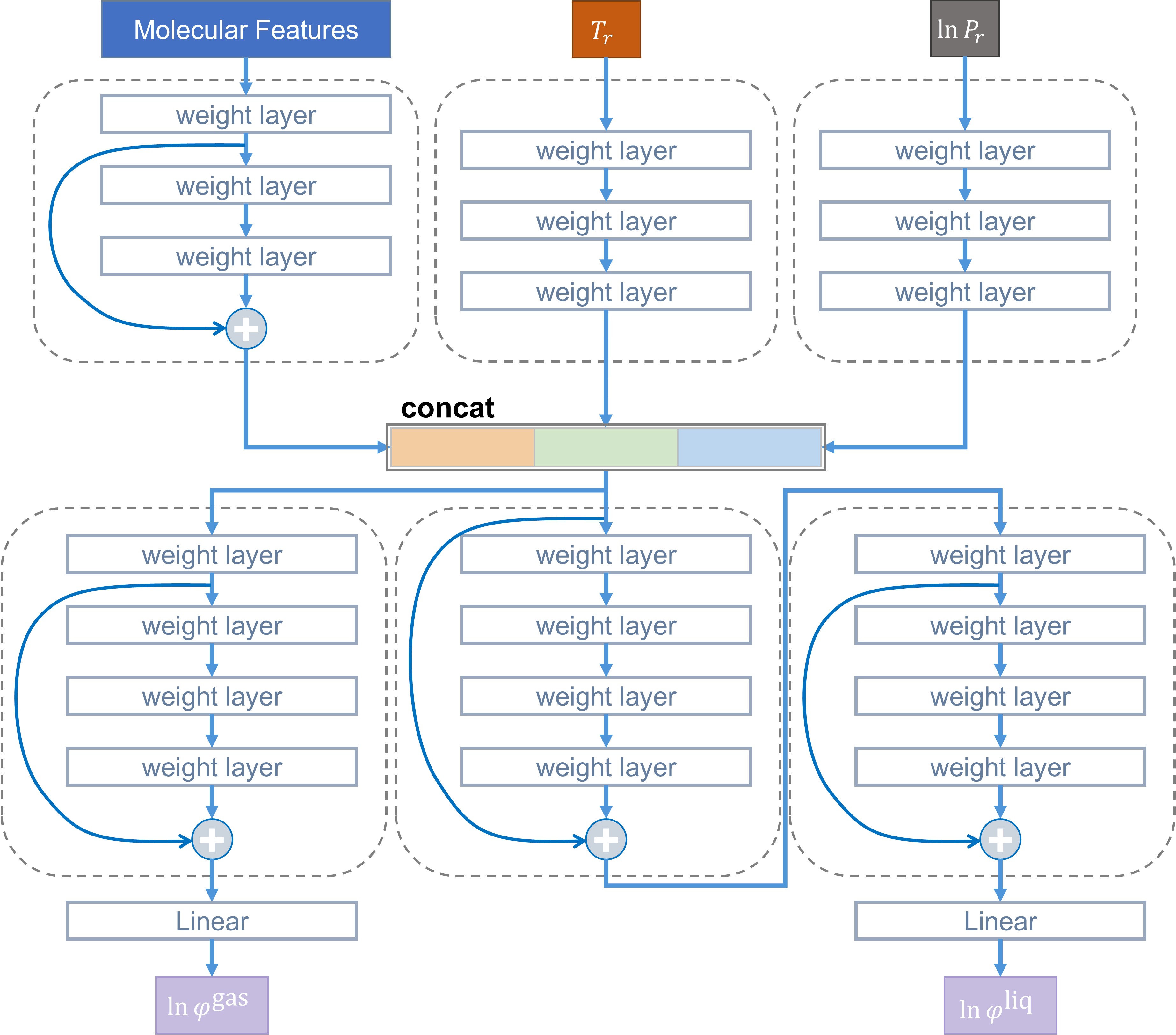}
    \caption{
    Model architecture diagram.
    The ``weight layer'' contains a linear layer and an activation function. ``concat'' denotes concatenating the three feature vectors together.
    }
    \label{fig:Network}
\end{figure}

The model takes molecular features, $T_r$, and $\ln p_r$ as inputs because they are the natural variables of Gibbs free energy. After being embedded by neural networks and concatenated into a single feature vector, this vector is fed into a residual neural network to compute $\ln \varphi^{\text{gas}}$ and $\ln \varphi^{\text{liq}}$ (FIG.~\ref{fig:Network}).

The model predicts fugacity coefficients for both the gas and liquid phases, which in practice are expressed as:
\begin{equation}
\ln \varphi=\left\{
    \begin{aligned}
        &\ln \varphi^{\text{liq}} &\text{if } p \geq p_c \text{ and } T < T_c \\
        &\ln \varphi^{\text{gas}} &\text{if } T \geq T_c \\
        &\min\{\ln \varphi^{\text{liq}}, \ln \varphi^{\text{gas}}\} &\text{others}
    \end{aligned}
\right.
.
\end{equation}
This calculation method treats the entire supercritical region as a gas phase. Below the critical point, it selects the phase with the smaller calculated fugacity coefficient between the gas and liquid phases, that is, the phase with a lower Gibbs free energy and greater thermodynamic stability.

\subsubsection{Loss Function}
The GEoS model training consists of two stages: pre-training and fine-tuning. 
Their primary difference lies in the loss function design.
The pre-training stage aims to capture the overall trend of $\ln \varphi$.
The fine-tuning stage focuses on the behavior of the small-magnitude $Z$. 

During pre-training, all loss function terms are dimensionless to ensure consistent magnitudes. 
The $\ln \varphi$ serves as the primary component of the loss function, where the model mainly learns the $\ln \varphi$ under various temperature and pressure conditions:
\begin{equation}
    \mathcal{L}_{\text{fug}} = \frac{1}{N}\sum^N_{i=1}\text{smoothL1}\left(\ln \varphi_i, \ln \hat{\varphi}_i\right)
\end{equation}
where
\begin{equation}
    \text{smoothL1}(x, y) = 
    \begin{cases} 
        0.5 (x - y)^2, & \text{if } |x - y| < 1 \\
        |x - y| - 0.5, & \text{otherwise}
    \end{cases}
    .
\end{equation}

In addition, we add a constraint term for the $Z$:
\begin{equation}
    \mathcal{L}_Z = \frac{1}{N}\sum^N_{i=1}\text{smoothL1}\left(Z_i, \hat{Z}_i\right)
\end{equation}

Considering that the $Z$ cannot be negative, we add a penalty term for negative values:
\begin{equation}
    \mathcal{L}_{Z>0} = \sum^N_{i=1} \max \left\{0, -Z_i\right\}
\end{equation}

Finally, we consider the mechanical stability constraint:
\begin{equation}
    \mathcal{L}_{\text{st}} = \sum^N_{i=1} \max \left\{0, p_i\left(\frac{\partial Z_i}{\partial p_i}\right)_{T_i} - Z_i\right\}
\end{equation}

The final pre-training loss function is:
\begin{equation}
    \mathcal{L} = \mathcal{L}_{\text{fug}} + \alpha \mathcal{L}_Z + \beta \mathcal{L}_{Z>0} + \gamma \mathcal{L}_{\text{st}} + \omega \mathcal{L}_{\text{L2}}
\end{equation}

Where $\alpha$, $\beta$, and $\gamma$ are scaling coefficients estimated from the magnitude differences between $\mathcal{L}_{\text{fug}}$ and other loss terms, aiming to align the scales of all loss terms.
$\mathcal{L}_{\text{L2}}$ is the L2 regularization term, and $\omega$ is set to $10^{-3}$.

During the fine-tuning phase, considering that the value of $Z$ spans multiple orders of magnitude and is difficult for the model to learn when $Z<10^{-2}$, the loss term for the compressibility factor is modified as:
\begin{equation}
    \mathcal{L}_Z = \frac{1}{N}\sum^N_{i=1}\text{ASinhLoss}\left(Z_i, \hat{Z}_i\right)
\end{equation}
where
\begin{equation}
    \text{ASinhLoss}\left(Z, \hat{Z}\right) = 
    \begin{cases}
        0 & \text{if } Z \leq \tau \\ 
        \text{smoothL1}\left(\text{ASinh}\left(Z/\varepsilon\right), \text{ASinh}\left(Z_i/\varepsilon\right)\right) & \text{if } Z > \tau
    \end{cases}.
\end{equation}
The threshold $\tau$ is set to $10^{-6}$, meaning  the samples of $Z \leq 10^{-6}$ are excluded from fine-tuning. At this magnitude, values are extremely small and approach the effective precision of single-precision floating-point format. 
The scaling factor $\varepsilon$ is set to $10^{-4}$. This scaling ensures that terms with $Z = 10^{-5}$ fall into the approximately linear region of the ArcSinh function, while larger terms fall into its approximately logarithmic region.

Additionally, during the fine-tuning stage, we observe that an overly strong mechanical stability constraint renders the compression factor constraint ineffective. Consequently, the mechanical stability constraint in fine-tuning is scaled based on pressure:
\begin{equation}
    \mathcal{L}_{\text{st}} = \sum^N_{i=1} \max \left\{0, \left(\frac{\partial Z_i}{\partial p_i}\right)_{T_i} - \frac{Z_i}{p_i}\right\}
\end{equation}

For more details on the design of loss function and training processes, please refer to \textit{Section 3.1} and \textit{Section 3.2} of the \textit{Supporting Information}.

\subsection{Evaluation Metrics}
Let $y_i$ denote the experimental value and $\hat{y}_i$ denote the model predicted value.
Let $\bar{y}$ denote the average of the experimental data:
\begin{equation}
    \bar{y} = \frac{1}{N} \sum_{i=1}^{N} y_i
\end{equation}
where $N$ is the number of experimental data points.

To evaluate the prediction performance of the model, the following metrics are adopted:
\begin{itemize}
    \item[1)] Root Mean Square Error (RMSE):
    \begin{equation}
        \text{RMSE} = \sqrt{\frac{1}{N} \sum_{i=1}^{N} \left( \hat{y}_i - y_i \right)^2}
    \end{equation}
    \item[2)] Mean Absolute Error (MAE):
    \begin{equation}
        \text{MAE} = \frac{1}{N} \sum_{i=1}^{N} \left| \hat{y}_i - y_i \right|
    \end{equation}
    \item[3)] Average Absolute Relative Deviation (AARD):
    \begin{equation}
        \text{AARD} = \frac{1}{N} \sum_{i=1}^{N} \left| \frac{\hat{y}_i - y_i}{y_i} \right| \times 100\%
    \end{equation}
    \item[4)] Coefficient of Determination ($R^2$):
    \begin{equation}
        R^2 = 1 - \frac{\sum_{i=1}^{N} \left( \hat{y}_i - y_i \right)^2}{\sum_{i=1}^{N} \left( y_i - \bar{y} \right)^2}
    \end{equation}
\end{itemize}

AARD mainly reflects the percentage deviation between predicted and true values and eliminates the influence of units. It is particularly suitable for physical quantities that span several orders of magnitude in chemical engineering, such as pressure and density, making deviations comparable across different levels of true values. In contrast, MAE retains the original units and directly describes the absolute magnitude of prediction errors, allowing direct comparison with engineering accuracy requirements. Therefore, this study uses both AARD and MAE as main evaluation metrics to account for the comparability of relative deviations and the interpretability of absolute deviations.
\section{Results and Discussions}
\subsection{Overall performance in the test set}
We collect a large number of density data from the ThermoData Engine (TDE) and construct an independent test set to evaluate the generalization performance of the model. The test set contains 74,078 liquid-phase density data points for 399 substances and 9,665 gas-phase density data points for 82 substances, none of which are included in model training. 

\begin{figure*}
    \centering
    \includegraphics[width=\linewidth]{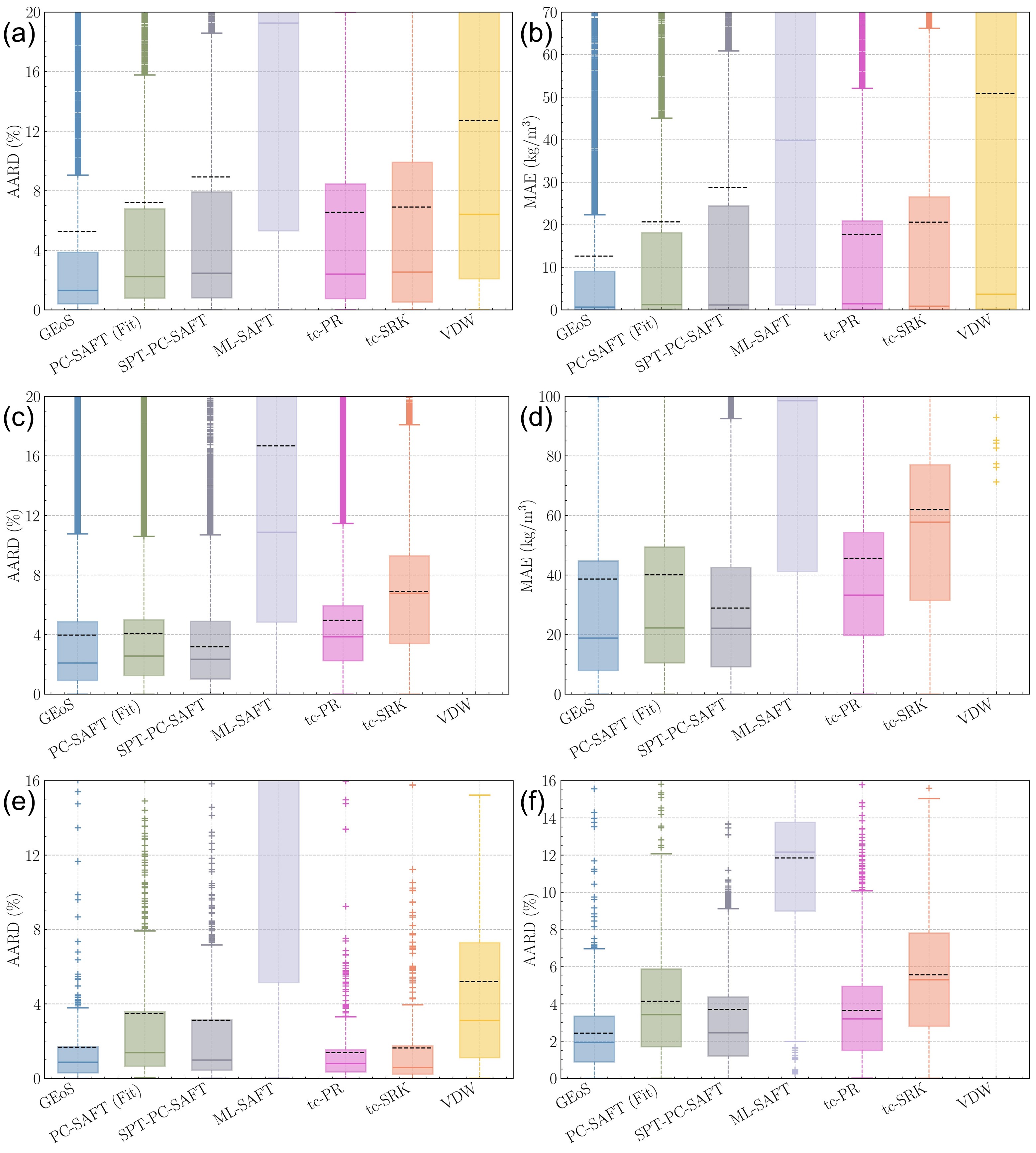}
    \caption{
    The predictive performance of different models with different metrics on the test set, where the central solid line is the median line, the upper and lower boxes are the upper and lower quartile lines, and the dash line is the mean line.
    (a, b) the prediction errors of the models for the gas-phase density;
    (c, d) the prediction errors of the models for the liquid-phase density;
    (e) the prediction errors of the gas-phase density for substances containing elements ($\ce{Br}$) outside the training set;
    (f) the prediction errors of the liquid-phase density for substances containing elements ($\ce{Br}$) outside the training set;
    Note that, in some cases, the errors of the van der Waals equation and ML-SAFT are so large that their corresponding box plots may extend beyond the range shown in the figure.
    }
    \label{fig:Capabilities}
\end{figure*}

We select two categories of models for comparison.
The first category comprises publicly available state-of-the-art machine learning models from existing studies, namely SPT-PC-SAFT\cite{Winter2025} and ML-SAFT\cite{Felton2024}, both of which predict PC-SAFT parameters.
The second category consists of commonly used cubic equations of state, including the translated-consistent Peng--Robinson (tc-PR)\cite{LeGuennec2016}, the translated–consistent Soave--Redlich--Kwong (tc-SRK)\cite{LeGuennec2016}, and the van der Waals (VDW) equation. However, since neither DIPPR nor TDE provides acentric factor data for some substances, tc-PR and tc-SRK cannot be applied to these substances. Accordingly, the evaluation scope of the three benchmark equations of state is reduced to 66 substances with 8,019 data points for the gas phase, and 362 substances with 68,415 data points for the liquid phase. All other benchmark models are evaluated on the complete test set.
The detailed performance metrics are presented in Table~\ref{tab:result} and FIG.~\ref{fig:Capabilities}.

\begin{table*}[!ht]
    \centering
    \caption{
    The predictive performance of each model on the test set; for AARD and MAE, the value before the parentheses is the mean, and the value in the parentheses is the median.
    }
    \label{tab:result}
    \begin{tabular}{llllll}
    \toprule
         & Model & AARD(\%) & MAE($\unit{kg/m^3}$) & RMSE($\unit{kg/m^3}$) & $R^2$ \\
    \midrule
        Gas & GEoS & \textbf{5.25}(\textbf{1.30}) & \textbf{12.63}(\textbf{0.64}) & \textbf{38.14}  & \textbf{0.9849}  \\
        ~ & PC-SAFT(Fit) & 7.22(2.24) & 20.66(1.21) & 48.84  & 0.9753  \\
        ~ & SPT-PC-SAFT  & 8.92(2.47) & 28.73(1.17) & 72.40  & 0.9457  \\
        ~ & ML-SAFT  & 404.20(19.26) & 239.82(39.83) & 418.31  & -0.8129  \\
        ~ & tc-PR  & 6.52(2.40) & 18.96(1.43) & 39.36  & 0.9818  \\
        ~ & tc-SRK  & 6.93(2.54) & 22.07(0.85) & 44.95  & 0.9762  \\
        ~ & VDW  & 13.16(6.41) & 54.65(3.66) & 100.28  & 0.8815 \\
        \midrule
        Liquid & GEoS & 3.96(\textbf{2.09}) & 38.62(\textbf{18.85}) & 79.22  & 0.9019  \\
        ~ & PC-SAFT(Fit) & 4.07(2.56) & 40.03(22.23) & 75.48  & 0.9109  \\
        ~ & SPT-PC-SAFT  & \textbf{3.18}(2.34) & \textbf{28.91}(22.15) & \textbf{41.15}  & \textbf{0.9735}  \\
        ~ & ML-SAFT & 16.67(10.87) & 159.26(98.51) & 274.49  & -0.1788  \\
        ~ & tc-PR  & 4.95(3.86) & 45.6(33.22) & 65.69  & 0.9259  \\
        ~ & tc-SRK  & 6.89(6.78) & 61.93(57.76) & 78.70  & 0.8937  \\
        ~ & VDW  & 41.49(41.22) & 381.03(358.26) & 396.12  & -1.6940  \\  
    \bottomrule
    \end{tabular}
\end{table*}

First, the results in Table~\ref{tab:result} show that the ML-SAFT and VDW models have poor predictive performance. In some cases, $R^2 < 0$, indicating limited predictive ability on the current test set. The results in FIG.~\ref{fig:Capabilities} further show that the predictions of these two models contain many substantial deviations, which are not caused by a few individual outliers.

As shown in FIG.~\ref{fig:Capabilities}(a, b), GEoS exhibits a significant advantage in gas-phase density prediction. 
According to Table~\ref{tab:result}, GEoS achieves an AARD of only 5.25\% and an MAE of \qty{12.63}{kg/m^3}. These values are much lower than those of the existing SPT-PC-SAFT method (8.92\% and \qty{28.73}{kg/m^3}) and also outperform tc-PR (6.52\% and \qty{18.96}{kg/m^3}). 
For liquid-phase density prediction, GEoS also performs well with an AARD of approximately 3.96\%, which is slightly worse than the 3.18\% of SPT-PC-SAFT; however, its median AARD is only 2.09\%, lower than the 2.34\% of SPT-PC-SAFT.
As shown in FIG.~\ref{fig:Capabilities}(c, d), the upper quartile of GEoS is largely consistent with that of SPT-PC-SAFT, but significantly lower than those of tc-PR and tc-SRK.
This indicates that the overall prediction accuracy of the proposed model is comparable to current machine learning methods, while being significantly superior to cubic equations of state.
The performance of MAE is similar to that of AARD. The median values of AARD and MAE are better while their mean values are worse, indicating that extreme deviation points have a large impact on the average metrics of GEoS.
Note that the SPT-PC-SAFT training set contains about 12,019 molecules, which is much larger than the 101 molecules in our training set. Coincidentally, all molecules in our test set are included in its training set. Thus, the current evaluation scheme favors SPT-PC-SAFT.

Furthermore, we evaluate the generalization ability of the model on compounds containing an atomic type not covered by the training set (i.e., \ce{Br}) to examine its prediction robustness for unknown chemical compositions. As shown in FIG.~\ref{fig:Capabilities}(e, f), box plots for \ce{Br}-containing substances are presented separately. With these two advantages, GEoS still outperforms existing models on substances containing atoms absent from the training set. These results indicate that GEoS has stronger extrapolation ability and higher accuracy, demonstrating the soundness of its feature engineering and model design.

Interestingly, the PC-SAFT model fitted from GEoS sampling also shows excellent performance on the test set. It outperforms the SPT-PC-SAFT model in gas density prediction and the tc-PR model in liquid density prediction, with almost no loss in prediction accuracy compared to the original GEoS.
Compared with GEoS, the fitted PC-SAFT model exhibits superior thermodynamic consistency in the low-pressure liquid-phase region and the high-pressure region.
It also enables efficient calculation of various thermodynamic properties, which reduces computational cost and facilitates integration into process design and simulation.

\subsection{Discussion of UMAP results}
\begin{figure*}[!htb]
    \centering
    \includegraphics[width=\textwidth]{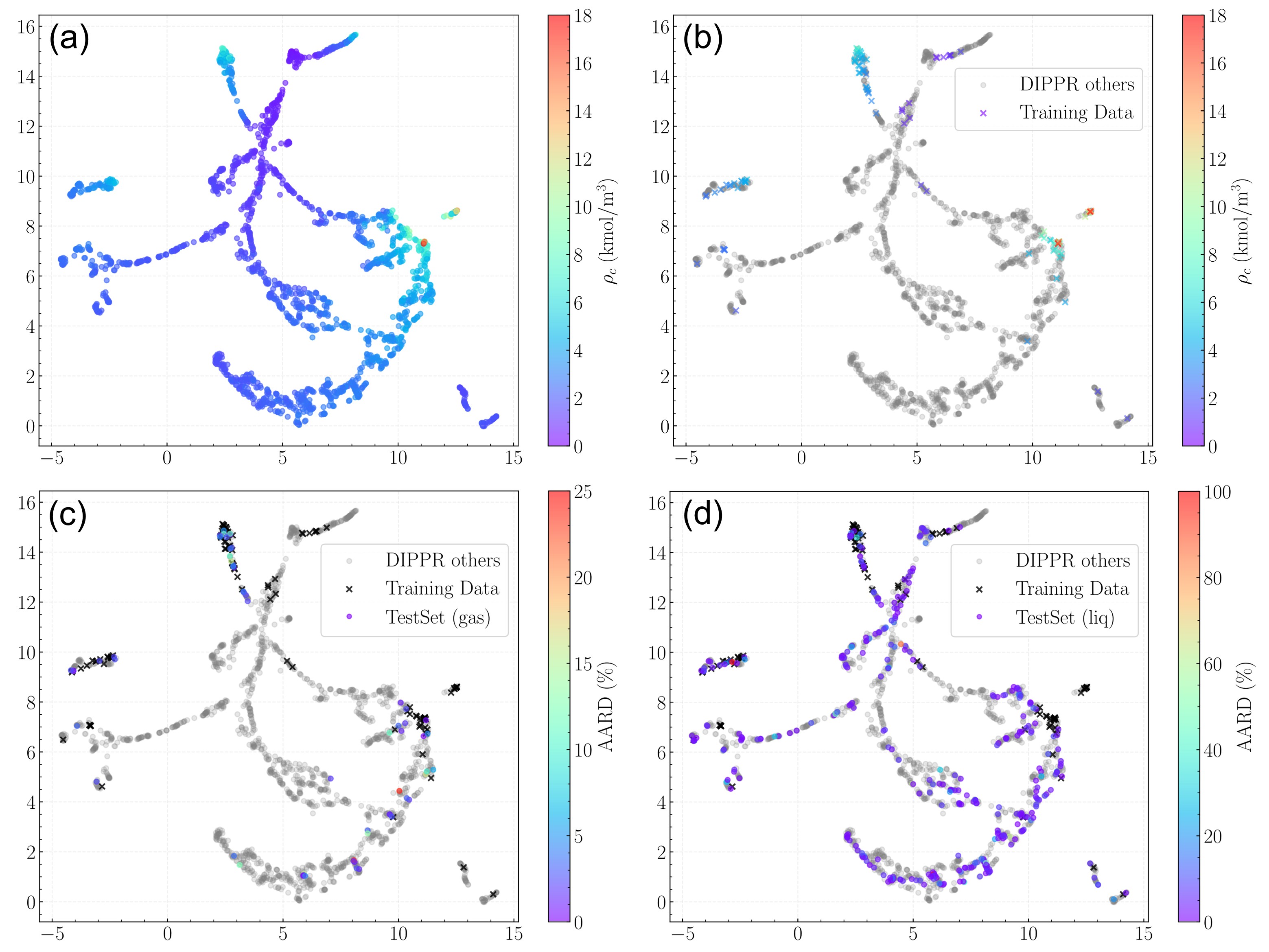}
    \caption{
    Distribution of DIPPR molecular features after UMAP reduction.
    (a) Data points colored by $\rho_c$ values;
    (b) Training set data colored by $\rho_c$ values, with other molecules colored gray;
    (c) Gas-phase density test set points colored by prediction AARD per compound;
    (d) Liquid-phase density test set points colored by prediction AARD per compound.
    }
    \label{fig:umap}
\end{figure*}

We apply the UMAP algorithm \cite{UMAP2020} to reduce the high-dimensional molecular features of the entire DIPPR database into two dimensions, as shown in FIG.~\ref{fig:umap}.

FIG.~\ref{fig:umap}(b) shows that, since the training set contains only 101 molecules, they occupy very small clusters. 
According to the results in FIG.~\ref{fig:umap}(c), the molecules in the gas-phase density test set are mainly distributed near the training set, with a few molecules located in the lower region. Overall, the performance is good.
FIG.~\ref{fig:umap}(d) shows the excellent generalization ability of the model: the samples in the liquid-phase density test set are distributed almost uniformly across the clusters. Even in regions where no training molecules appear, the model still maintains very high prediction accuracy, except for a few points.

The UMAP results fully demonstrate the potential of the GEoS model. 
Despite being trained on a small-sample dataset, the model maintains high prediction accuracy in out-of-distribution regions. This not only confirms its excellent generalization capability but also indicates that it learns the underlying laws of the EoS, promising to be an efficient machine learning paradigm for fluid EoS modeling.

\subsection{Discussion of Intermolecular Potentials}
From the perspective of the Lennard-Jones (LJ) potential, we analyze the underlying laws of the EoS.
Since it is mathematically difficult to inversely derive the LJ potential from the EoS, we use GEoS sampling and fit the data using the SAFT-VR-Mie EoS \cite{Lafitte2013} with $m=1$ provided by $\mathrm{FeO_s}$ package\cite{rehner2023feos}. 

\begin{figure*}
    \centering
    \includegraphics[width=\linewidth]{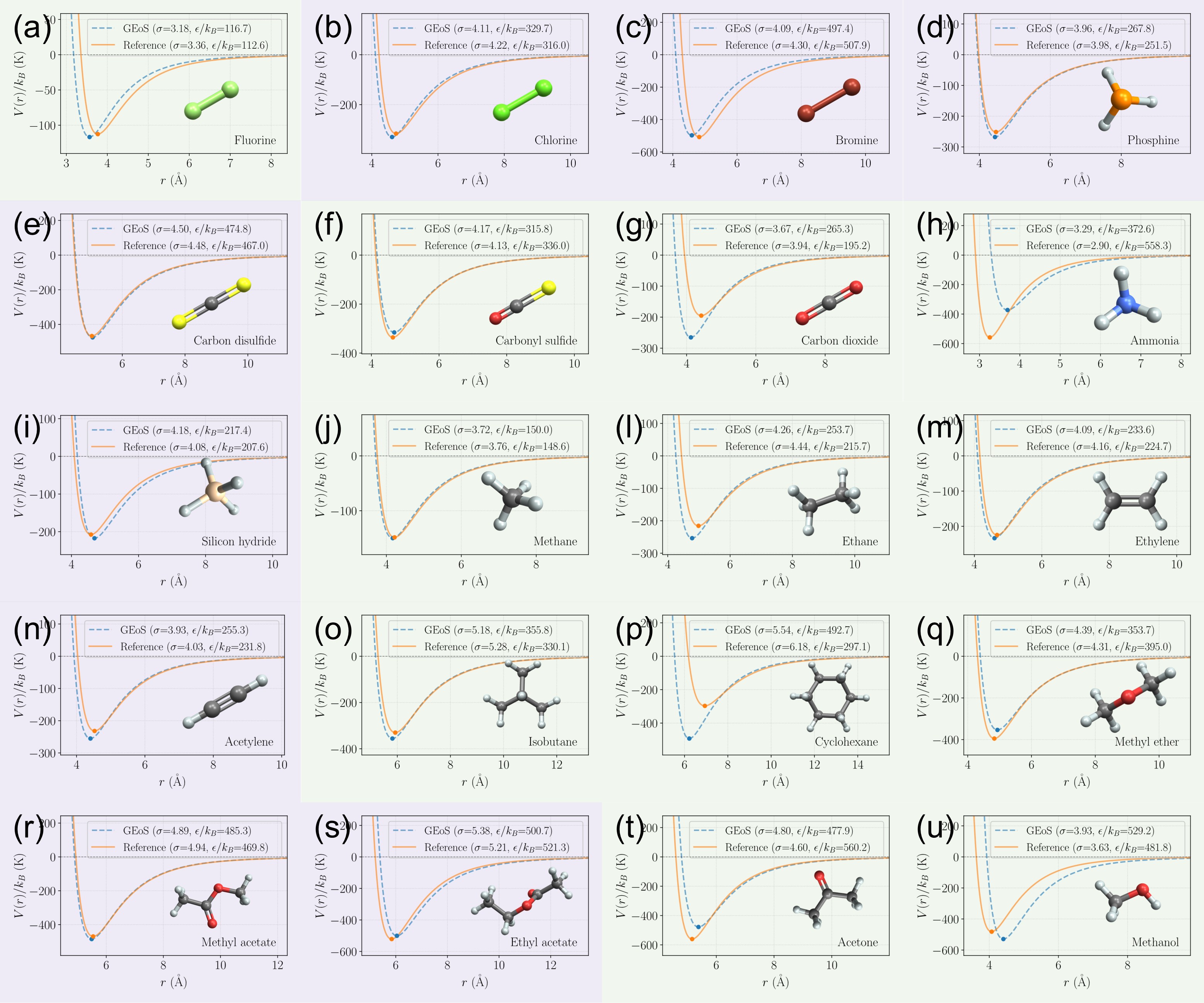}
    \caption{
    Comparison between the fitting results based on GEoS sampling and the reference Lennard-Jones potentials~\cite{elliott2023properties}.
    Molecules with a green background are in the training set, while those in purple are not.
    }
    \label{fig:LJ}
\end{figure*}

The reference molecular Lennard-Jones potentials are taken from the book by Elliott et al.~\cite{elliott2023properties}.
The test results show that the prediction error of $\sigma$ is about 4.99\% and the error of $\epsilon/k_B$ is about 21.44\% over 54 small molecules.
We select some typical substances and plot their potential energy curves, as shown in FIG.~\ref{fig:LJ}.

The LJ potentials of most molecules closely match the reference results. 
This is likely because these molecules have lower polarity and are more spherical. 
In contrast, some molecules are highly polar, and some (e.g., $\ce{NH3}$ and $\ce{CH3OH}$) exhibit hydrogen bonding. A simple LJ potential is insufficient for these cases, leading to larger fitting deviations.

The results indicate that GEoS essentially captures the intermolecular interaction patterns well. We believe this serves as a crucial foundation for the model's strong generalizability and its ability to accurately predict gas and liquid phase densities for novel substances.

\section{Conclusions and Outlook}
This study develops a fluid equation-of-state model, GEoS. The model uses predefined 2D and 3D molecular descriptors to predict fluid fugacity coefficients through thermodynamically consistent neural networks. The PVT relations of the fluids are then obtained using automatic differentiation.
This study is among the few efforts to construct a fluid equation of state from the perspective of Gibbs energy. It is expected to provide a new perspective for the future development of fluid equations of state.

The GEoS training set contains only 101 molecules. However, when evaluated on a test set comprising 74,078 liquid-phase density data points for 399 entirely new substances and 9,665 gas-phase density data points for 82 entirely new substances, GEoS shows excellent performance. It outperforms the current volume-translated SRK and PR equations, as well as the SPT-PC-SAFT model trained on 13,279 molecules.
Furthermore, through UMAP analysis, we demonstrate the strong generalization ability of GEoS. By fitting the Lennard-Jones (LJ) potential, we also validate the hypothesis that the model learns the underlying principles and physical mechanisms of equations of state.

In our tests, we find that the GEoS model exhibits certain anomalies in its predictions for the low-pressure liquid phase and the ultra-high-pressure region. Since the PC-SAFT equation of state itself offers high accuracy and good thermodynamic consistency, we propose using GEoS to fit the PC-SAFT parameters, so as to eliminate the above anomalies while maintaining accuracy. In addition, the fitted PC-SAFT equation lays a foundation for its subsequent implementation in process simulation software and engineering applications.

In future work, we plan to expand the training set to develop a more general equation of state prediction model.
Additionally, while the current model predicts only the pure component $\ln \varphi$, we aim to extend it to predict the partial fugacity coefficient of mixtures, $\ln \hat{\varphi}_i$.
Finally, we will leverage the findings from our neuron activation analysis to guide model design, aiming to establish a novel modeling approach that bridges neural network structural analysis with structural design.

\begin{acknowledgments}
We thank Yeou Han, Zhenning Guo, Ang Ji, Shangkun Shi, Zhengye Lu, Xingyu Zhong, and Yifei Chen for their preliminary exploration in compression factor modeling.
\end{acknowledgments}

\section*{DATA AVAILABILITY}
The dataset construction code, training code, fitted PC-SAFT parameters, and test set predictions of all models in this study are fully available on \href{https://github.com/A-Normal-User/GEoS}{Github}.
Due to licensing restrictions, the critical point information in the test set, sourced from the DIPPR 801 database, cannot be made publicly available. Similarly, the density data in the test set originate from the NIST ThermoData Engine and are also not publicly shared.
Interested readers may access \href{https://app.knovel.com/kn/resources/kpDIPPRPF7/toc}{Knovel} to obtain the critical point data and use the \href{https://trc.nist.gov/tde.html}{ThermoData Engine} to acquire the density values for the test set.

\nocite{*}

\bibliography{ref/refs}% Produces the bibliography via BibTeX.

\end{document}